\documentclass[letterpaper,10pt]{article} 
\usepackage[utf8]{inputenc}
\usepackage{graphicx}
\usepackage{subcaption}
\usepackage{amsmath, amssymb}
\usepackage{authblk}  % Author formatting
\usepackage{cite}     % Sorted/compressed citations
\usepackage{geometry}
\usepackage[toc,page]{appendix}
\usepackage[colorlinks=true,bookmarks=false,citecolor=blue,urlcolor=blue]{hyperref}
\usepackage{cleveref}
\usepackage{titlesec}
\usepackage{xcolor}
\usepackage{tocloft}

\titleformat{\section}
  {\bfseries\Large}{\thesection}{1em}{}

\titleformat{\subsection}
  {\bfseries\large}{\thesubsection}{1em}{}

\titleformat{\subsubsection}
  {\bfseries\normalsize}{\thesubsubsection}{1em}{}

\title{Demystifying Flux Architecture}

\author[1,2]{Or Greenberg}
\affil[1]{Hebrew University of Jerusalem, Israel}
\affil[2]{General Motors R\&D, SDVR}

\date{}  % Suppress date

\begin{document}

\maketitle

\begin{figure}[hbtp]
    \centering
    \includegraphics[width=0.6\textwidth]{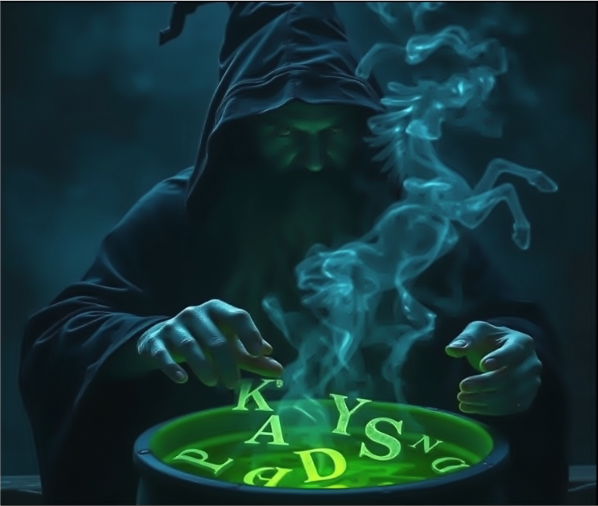}
    \caption{Teaser image.}  % Optional: add a caption for accessibility
    \label{fig:teaser}
\end{figure}

\begin{abstract}
FLUX.1 is a diffusion-based text-to-image generation model developed by Black Forest Labs, designed to achieve faithful text-image alignment while maintaining high image quality and diversity. FLUX is considered state-of-the-art in text-to-image generation, outperforming popular models such as Midjourney, DALL·E 3, Stable Diffusion 3 (SD3), and SDXL. Although publicly available as open source, the authors have not released official technical documentation detailing the model’s architecture or training setup. This report summarizes an extensive reverse-engineering effort aimed at demystifying FLUX’s architecture directly from its source code, to support its adoption as a backbone for future research and development. This document is an \textit{unofficial} technical report and is \textbf{not published or endorsed by the original developers or their affiliated institutions}.  
\end{abstract}

\newpage
\tableofcontents

\newpage
\section{Background}

To better understand the foundations of FLUX, we briefly present several key models and methods that preceded it and influenced its design. This overview is intentionally high-level, aiming only to provide the essential background necessary to grasp the core principles behind FLUX. For readers interested in a deeper exploration of these foundational works, references to the original papers and technical reports are provided throughout this section.

\subsection{ML-Based Image Generation}
The task of ML-based image generation refers to a family of models designed to generate novel samples from a simple distribution that emulates a source dataset. Specifically, some models learn to map a dataset of images to a simple distribution (e.g., Gaussian), from which new images can be sampled. Several types of generative models have been developed in recent years, including—but not limited to—Variational Autoencoders (VAEs) \cite{kingma2013auto}, Generative Adversarial Networks (GANs) \cite{goodfellow2020generative}, Normalizing Flows (NFs) \cite{papamakarios2021normalizing}, and Diffusion Models (DMs) \cite{ho2020denoising}.

Among these, text-to-image models (e.g., DALL·E \cite{ramesh2021zero}, Stable Diffusion \cite{rombach2022high}, FLUX \cite{flux2024}) have achieved state-of-the-art performance and are widely used in applications that generate high-quality images from textual instructions, commonly referred to as “prompts.”

\subsection{Diffusion Models}
\label{subsec:DM}

Within this scope, Diffusion Models (DMs) \cite{ho2020denoising} have recently emerged as the state-of-the-art (SoTA) approach for text-conditioned image generation. They work by learning to reverse a gradual noising process applied to training data. While early diffusion models relied on U-Net architectures composed of convolutional layers augmented with attention mechanisms for image–text alignment, recent models such as SD3 \cite{esser2024scaling} and FLUX.1 \cite{flux2024} have transitioned to fully Transformer-based architectures for the denoising process, offering improved scalability and context modeling. During the training process, clean images are incrementally corrupted by Gaussian noise whose magnitude is determined by a timestep-specific schedule, and the model learns to iteratively denoise them. 

In the most common version of DM optimization, the model is trained to predict the normalized added noise $\epsilon$, and optimized using a reconstruction loss as formulated in \Cref{eq:DM}

\begin{equation}\label{eq:DM}
    \mathcal{L}_\epsilon = \mathbb{E}_{x_t,\epsilon,t}
    \left[  \|\epsilon_\theta(x_t,t) -\epsilon\|^2  \right]
\end{equation}

where $x_t = \sqrt{\alpha_t}x +\sqrt{1-\alpha_t}\epsilon$ represent the noisy version of the clean image $x_0$, corrupted by the gaussian noise $\epsilon \sim \mathcal{N}(0,1)$ to timestep $t$. \\

At inference time, the model begins from pure Gaussian noise and iteratively denoises it to produce realistic images. Unlike earlier generative models such as GANs, which rely on adversarial training, diffusion models optimize a reconstruction-based objective. This leads to more stable training and superior output quality, particularly for high-resolution image synthesis. Moreover, they offer fine-grained control over the generation process, enabling powerful applications such as super-resolution, inpainting, and text-conditioned image generation.

\subsection{Stable Diffusion}

Stable Diffusion is a family of latent diffusion models (LDMs) \cite{rombach2022high} designed to generate high-quality images conditioned on textual prompts. Unlike traditional pixel-space diffusion models, LDMs operate in a compressed latent space, significantly improving computational efficiency while maintaining visual fidelity. This is achieved by encoding input images into a lower-dimensional latent representation using a pre-trained autoencoder. The diffusion process is then applied in this latent space, and the final output is decoded back into pixel space.

The 1.x versions of Stable Diffusion (1.0 to 1.5) introduced this efficient framework for open-domain image synthesis, using a frozen CLIP \cite{radford2021learning} text encoder for prompt conditioning and a U-Net architecture for latent denoising. These models, trained on subsets of LAION-2B \cite{schuhmann2022laionb}, quickly gained popularity due to their open-source release, high versatility, and ease of fine-tuning.

Stable Diffusion v2.1 introduced several architectural upgrades, including a switch to OpenCLIP \cite{ilharco_gabriel_2021_5143773} and training on higher resolutions (768×768), improving image structure and fidelity. SDXL \cite{podell2023sdxl}, a major advancement in the series, uses a two-stage architecture with separate base and refiner models, allowing better handling of complex prompts and achieving significantly improved realism and prompt alignment. SD-Turbo \cite{sauer2024adversarial} further innovates by enabling near real-time image generation through a Adversarial-diffusion-distillation (ADD) process, where a student model trains to mimic the performance of a teacher model (SD2.1 or SDXL) using fewer inference steps.

Most recently, Stable Diffusion 3 (SD3) \cite{esser2024scaling} integrates a diffusion transformer backbone and adopts multimodal training (see \Cref{subsec:transformer} below), combining both text and image understanding for improved prompt adherence, compositional reasoning, and consistency. SD3 is designed for scalability and robustness, closing the gap between open models and proprietary systems like DALL·E 3 and Midjourney in terms of controllability and quality.

\subsection{Rectified Flows}
\label{subsec:RF}

Rectified Flow (RF) \cite{liu2022flow} is a recent generative modeling framework that simplifies and generalizes diffusion models by replacing stochastic denoising with a deterministic vector field, based on the Flow Matching principle introduced in \cite{lipman2022flow}. While RF is a training paradigm rather than an architectural change, its relevance to FLUX stems from the fact that FLUX was trained using this method. In this report, we do not provide a detailed explanation of Rectified Flow. Instead, we briefly highlight the key distinction between standard diffusion model training and the approach used in FLUX. This divergence in training schemes likely contributes to differences in performance and convergence behavior. For a comprehensive understanding of Rectified Flow, we refer readers to the original works.

While diffusion models (such as DDPMs \cite{ho2020denoising}) learn to predict the noise ($\epsilon$) added during a forward diffusion process (see \Cref{subsec:DM}), RFs train a model to predict a velocity vector $v$ that directly points from a noisy point $x_0$ (not to be confused with DM notation, where $x_0$ denotes the ``clean'' image) back to the data point $x_1$, along a straight interpolation path.

In diffusion models, the network is trained using the $\epsilon$-objective, which minimizes the difference between the predicted noise $\epsilon_\theta(x_t,t)$ and true noise $\epsilon$ added to a sample (see \Cref{eq:DM})

In contrast, RF defines a deterministic interpolation between the data $x_1$ and noise $x_0$, and trains the model $v_\theta$ to predict the velocity vector between the two:

\begin{equation}
    \mathcal{L}_v = \mathbb{E}_{x_0,x_1,t}
    \left[  \|v_\theta(x_t,t) -(x_1 - x_0)\|^2  \right]
\end{equation}

where $x_t=(1-t)x_0 + tx_1$ is a linear interpolation between $x_0 \sim \mathcal{N}(0, I)$ and $x_1 \sim p_{data}$, and $v(x_t, t)$ represents the target vector pointing from $x_0$ to $x_1$. 

Unlike diffusion models that rely on stochastic sampling, noise schedules, and denoising objectives, Rectified Flow leverages a velocity-based formulation that defines a deterministic transport path from noise to data. This approach eliminates the need for iterative noise perturbation and reverse denoising, enabling faster and more stable image synthesis. By solving a simple ODE with a learned velocity field, Rectified Flow achieves high-quality generation with significantly reduces complexity in both training and inference—making it a compelling and efficient alternative to traditional diffusion-based methods.

\subsection{Transformers}

Transformers are a class of neural network architectures originally introduced for natural language processing (NLP) \cite{vaswani2017attention}. The transformer’s core innovation is the self-attention mechanism, which enables the model to weigh relationships between all elements in an input sequence. This architecture has since become the foundation of many state-of-the-art models across modalities—including text, image, video, and multimodal systems.

In each attention block, the model computes three matrices from the input embeddings: queries (Q), keys (K), and values (V), using learned linear projections. The attention scores are computed as:

\begin{equation}
    Attention(Q, K, V) = softmax(\frac{QK^T}{\sqrt{d_k}}\cdot V)
\end{equation}

Here, $d_k$ is the dimensionality of the key vectors, and the softmax operation ensures the attention weights sum to 1. This mechanism allows the model to assign different levels of importance to different tokens when computing a new representation for each token in the sequence. In practice, multi-head attention is used, where multiple sets of $Q$/$K$/$V$ projections are computed in parallel to capture diverse types of relationships.

For vision tasks, Vision Transformers (ViTs) \cite{dosovitskiy2020image} tokenize an image into fixed-size non-overlapping patches (e.g., $16\times16$ pixels), flatten each patch, and project them into an embedding space. These patch embeddings are then processed by a standard transformer encoder, often with added positional encodings to preserve spatial structure. This enables ViTs to model long-range dependencies across an image without convolutional inductive biases.

While originally $Q$/$K$/$V$ represenations are all projected from the same input embedding in a process denotes \emph{Self-Attention}, In text-to-image generation, transformers often use \emph{Cross-Attention} to condition image synthesis on textual prompts. In this setup, the queries ($Q$) are projected from the image tokens, while the keys ($K$) and values ($V$) are projected from the text embeddings. This allows the model to modulate image generation based on semantic information from the prompt.
\section{FLUX.1}
\subsection{Introduction}

\textbf{FLUX.1} \cite{flux2024} is a rectified flow transformer trained in the latent space of an image encoder, introduced by \textbf{Black Forest Labs} in August 2024. 

\begin{figure}
\centering
\includegraphics[width=0.89\textwidth]{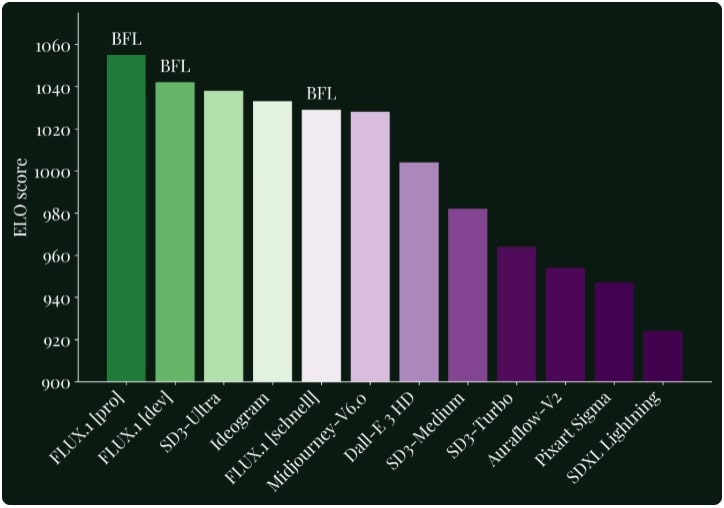}
  \caption{FLUX.1 defines a new state-of-the-art in image detail, prompt adherence, style diversity and scene complexity for text-to-image synthesis. Evaluation from \cite{FLUXAnnounce}}
\label{fig:ELO_score}
\end{figure}

\begin{figure}
\centering
\includegraphics[width=0.89\textwidth]{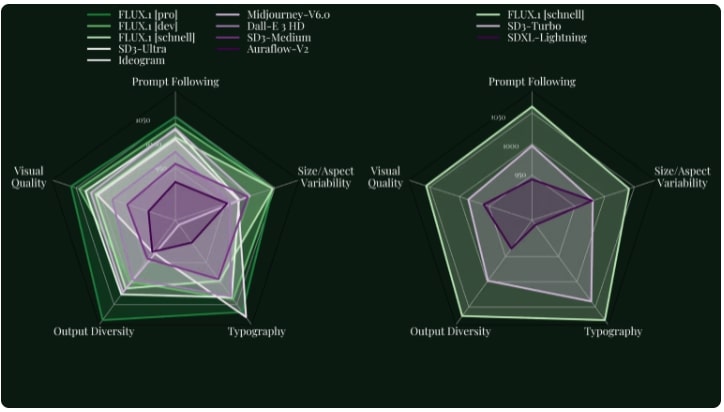}
  \caption{ELO scores for different aspects: Prompt Following, Size/Aspect Variability, Typography, Output Diversity, Visual Quality. Evaluation from \cite{FLUXAnnounce}}
\label{fig:ELO_score2}
\end{figure}

The FLUX.1 models (see \Cref{sec:hub}) demonstrate State-of-the-art (SoTA) performance for text-to-image tasks, in both terms of output quality and image-text alignment, as demonstrated in \Cref{fig:ELO_score,fig:ELO_score2} using the ELO-score metric, which ranks image generation models based on human preferences in head-to-head comparisons. A qualitative illustration and comparison to SD is provided in \Cref{Appendix}.

While the model adheres to the Rectified Flow training paradigm (according to the developers statement), the exact details regarding the training setup —including the dataset, scheduling strategy, and hyperparameters— have not been publicly disclosed. However, the model’s architecture and inference scheme can be reverse-engineered from the publicly available inference code. \\ 

\begin{figure}
\centering
\includegraphics[width=0.89\textwidth]{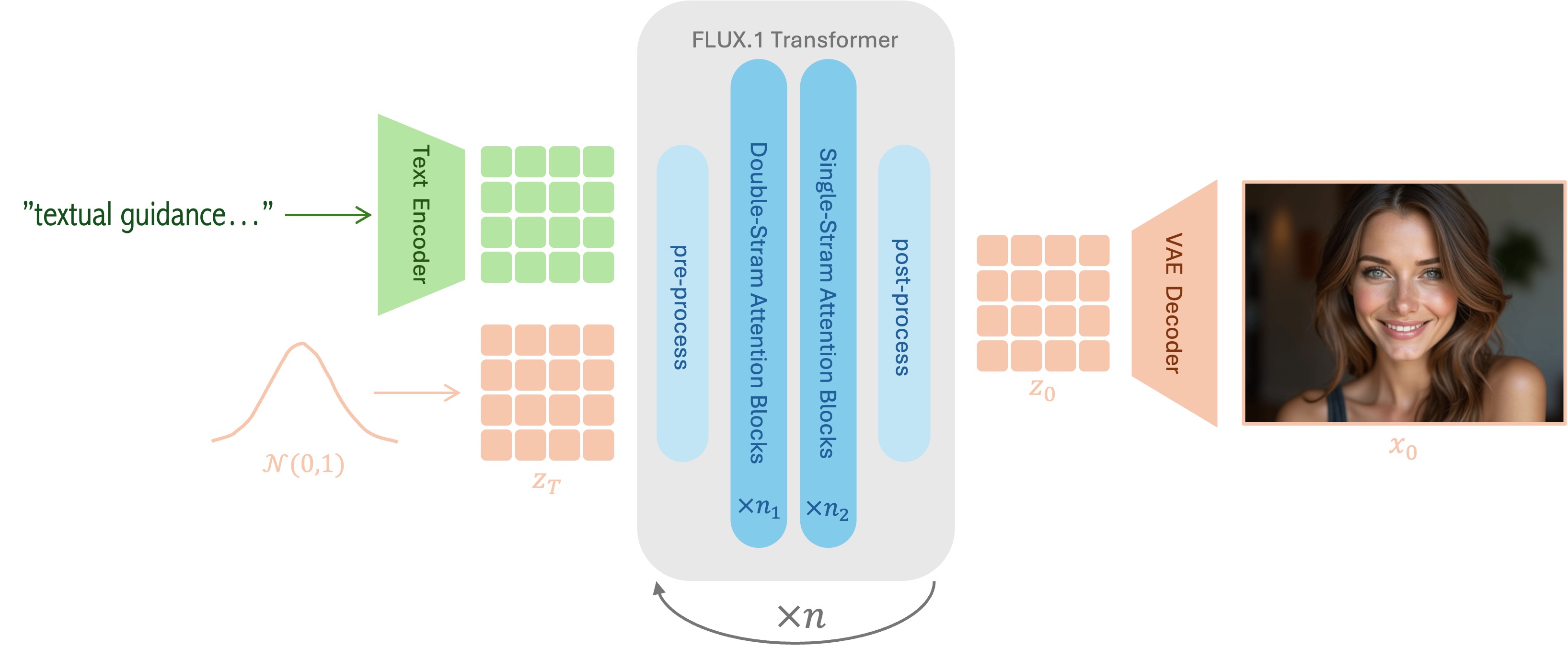}
  \caption{High-level overview of the FLUX.1 architecture. Text embeddings and latent image embeddings are iteratively processed through a series of attention blocks to generate a text-conditioned image.}
\label{fig:top-view}
\end{figure}

In this section, we outline the model’s architecture to demystify its behavior at inference time. A top-view of the architecture is illustrated in \Cref{fig:top-view}, where text embeddings and latent image embeddings are iteratively processed via a series of attention blocks. In the following section we deep-dive into the different components of the model. We begin with a high-level overview of FLUX’s sampling pipeline in \Cref{subsec:arch}, highlighting the key stages and the pre-trained components involved, followed by a deep-dive into the transformer's architecture in \Cref{subsec:transformer}, where a detailed explanation is provided to the different stages and concepts used to construct it.

\subsection{FLUX.1 Sampling Pipeline}
\label{subsec:arch}

In this section, we describe the sampling pipeline of FLUX.1. For simplicity, we refer to the text-to-image sampling process as being conditioned on a single prompt per sample.

Similar to LDM \cite{rombach2022high}, FLUX operates in a latent space, where the final latent output is decoded to reconstruct the RGB image in pixel space. Following LDM’s approach, the developers trained a convolutional autoencoder from scratch using an adversarial objective, but scaled up the latent representation from 4 channels (in LDM) to 16 channels.

\begin{figure}
\centering
\includegraphics[width=0.99\textwidth]{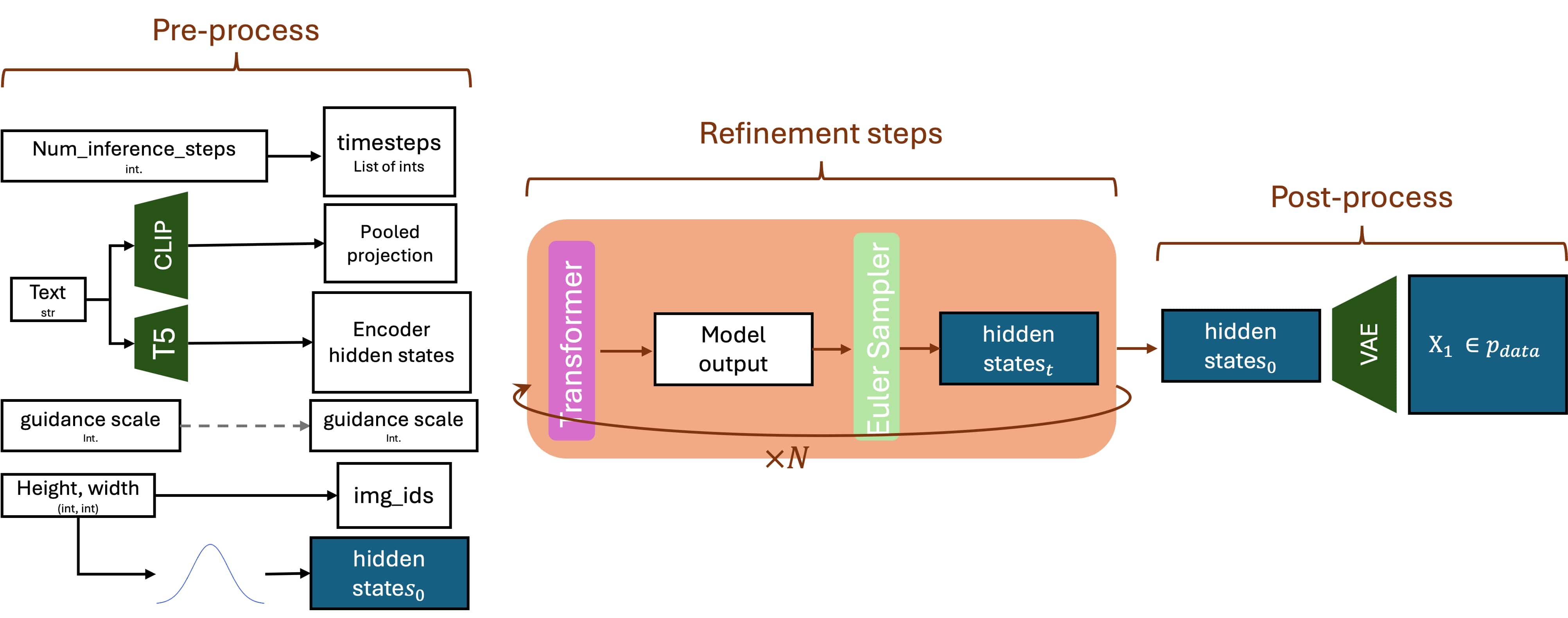}
  \caption{Flux.1 sampling pipeline. Just like in Diffusion Models, after pre-processing the noisy latent $z_t$ (denoted \emph{hidden\_state}$_t$) is being iteratively refined in the latent space. The final refined $z_0$ ($=$\emph{hidden\_states}$_0$) is decoded into an RGB image using a pre-trained VAE decoder.}
\label{fig:flux_pipeline}
\end{figure}

The sampling pipeline consists of three phases: $(1)$ \emph{Initiation and Pre-processing $(2)$} \emph{Iterative Refinement}, and $(3)$ \emph{Post-processing}. An overview of these steps is illustrated in \Cref{fig:flux_pipeline}, where the notations follows the ones used in the official implementation of FLUX.1 pipeline in diffusers \cite{von-platen-etal-2022-diffusers}. In the section below, we primarily focus on the \emph{Initiation and Pre-processing} phase of the pipeline. A brief overview  of the \emph{Iterative Refinement} and \emph{Post-processing} phases is provided at the end of this section. The \emph{Iterative Refinement} phase, in which the FLUX transformer operates, is discussed in detail in \Cref{subsec:transformer}.

\subsubsection{Initiation and Pre-processing}
\label{subsubsec:init}

In this phase, the model processes the user-provided inputs to prepare them for the main transformer’s refinement stage. A key component of this stage is the text encoder, which processes the textual guidance supplied by the user. FLUX.1 utilizes two pre-trained text encoders: \\

\noindent \textbf{CLIP Text Encoder.} 
CLIP (Contrastive Language–Image Pretraining) \cite{radford2021learning} is a foundation model developed by OpenAI, designed to bridge vision and language understanding. CLIP's text-encoder transforms natural language prompts into class level or dense, high-dimensional embeddings that can be directly compared with image embeddings in a shared latent space. Trained on hundreds of millions of image–text pairs, the CLIP text encoder captures rich semantic information, enabling models to align textual descriptions with corresponding visual content effectively, making it widely used in text-to-image generation tasks.
\\

\noindent \textbf{T5 Text Encoder.} 
T5 (Text-To-Text Transfer Transformer) \cite{raffel2020exploring} is a versatile language model developed by Google that frames all NLP tasks (e.g., translation, summarization, and question answering) as text-to-text problems. Its text encoder converts input text into contextualized embeddings using a Transformer-based architecture trained over massive language corpora. Unlike CLIP’s text encoder, which is trained jointly with an image encoder to produce embeddings aligned with visual features for contrastive learning, T5 is trained purely on textual data and optimized for language understanding and generation. This makes T5 well-suited for providing rich, token-level semantic representations even for long and complex textual prompmts.
\\

Below, we outline all the required inputs and describe how the model processes them in preparation for the refinement stage.

The pipeline's required inputs are:
\begin{itemize}
\item \textbf{text:} textual prompt guides the image generation process to enforce specified features or qualities.
\item \textbf{guidance\_scale:} controls the strength of conditioning.
\item \textbf{num\_inference\_steps:} how many iterative sampling steps should be performed.
\item \textbf{resolution:} Specifies the spatial resolution (height and width) of the generated image.
\end{itemize}

\vspace{1cm}

Once initiated with those inputs, they are being preprocessed as follows:

\begin{itemize}
\item \textbf{text:} The textual prompt is being encoded using two pre-trained text encoders:
\begin{itemize}
\item T5: provides dense (per token) embeddings, denoted \emph{encoder hidden states}
\item CLIP: provides pooled embeddings (one embedding for the whole prompt, like CLS embedding), denoted \emph{pooled projection}
\end{itemize}
\item \textbf{num\_inference\_steps:} Specifies the total number of sampling steps used during inference. It determines a subset of timesteps from the full diffusion range $t \in [0:T]$, where typically $t=1000$. Iterating over these selected timesteps defines the sampling trajectory.
\item \textbf{resolution:} The desired resolution determines the spatial dimensions of the initial (latent) noise sample $z_0 \sim \mathcal{N}(0,1)$. it is also used to define the img\_ids- a set of per-token indicators, pointing at the token's spatial location on a 2D grid. Given a target resolution (H,W) in pixel space, the corresponding latent dimensions $(h,w)$ are computed as $(h = H // VAE_{scale})$ and $(w = W // VAE_{scale})$ where $VAE_{scale}=8$. 
The image-token grid is further downsampled to dimensions $(h//2, w//2)$, and each token is assigned a unique identifier of the form $(t, \hat{h}, \hat{w})$, where $\hat{h} \in [0:h-1]$ and $\hat{w} \in [0,w-1]$, indicating the token’s location on a $2D$ spatial grid. text\_ids are initiated using the same structure of img\_ids, but with $t=h=w=0$ for all tokens. Formally, text\_ids = $n\cdot (0, 0, 0)$ where $n$ is the maximal number of tokens in T5 ($=512$).
\end{itemize}

\subsubsection{Iterative Refinement and Post-processing}

Let \emph{$\Phi =$ [encoder\_hidden\_states, pooled\_projection, guidance, img\_ids, text\_ids]} be a set the pre-processed pipeline's inputs. Same as in Diffusion Models, $[z_t, t,\Phi]$, are iteratively fed into the transformer during inference sampling, s.t.:

\begin{equation}
\forall\, t \in \text{timesteps}: \quad z_{t+\Delta t} = \text{Samp}(v_\theta(z_t, t, \Phi))
\end{equation}

Where $v_\theta$ is the trainable network that estimates the velocity vector (see \ref{subsec:RF}) and \emph{Samp($\cdot$)} refers to the Flow-Matching Euler Discrete sampler \cite{lipman2022flow}. Pay attention to the notation that differs from the one used in Diffusion Models. Here \emph{timesteps} ranges between $0$ to $1$, with $z_1$ the clear image and $z_0$ the pure Gaussian noise. In \Cref{subsec:transformer} we explore the architecture of $v_\theta$.

After the iterative refinement, the final clean latent $z_1$ is decoded via pre-trained VAE model to get the final image $x_1$. 

\subsection{Transformer}
\label{subsec:transformer}
\begin{figure}
\centering
\includegraphics[width=0.89\textwidth]{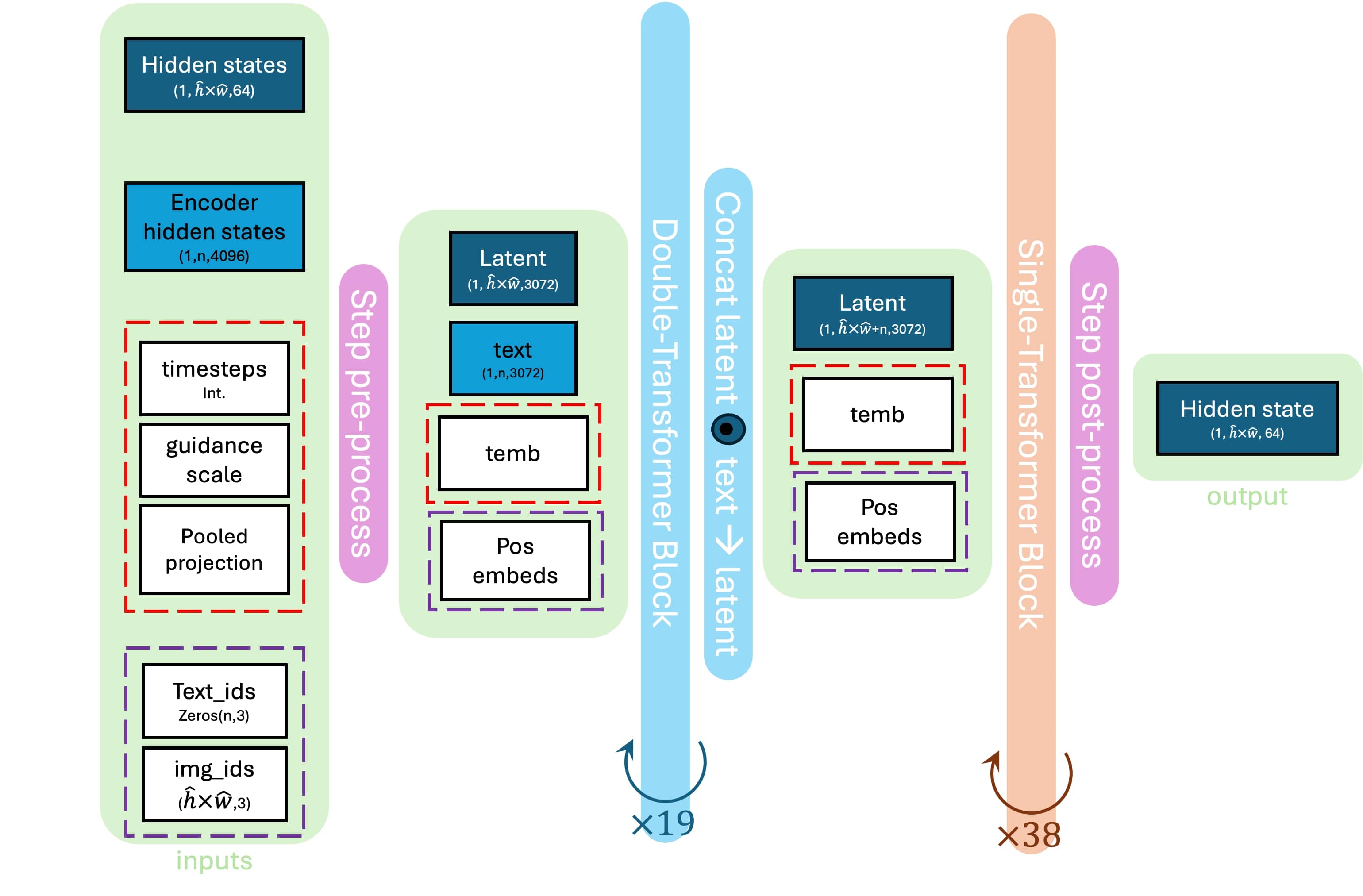}
  \caption{Flux Transformer: high-level overview}
\label{fig:transformer}
\end{figure}

The core component of FLUX.1’s synthesis pipeline is the velocity predictor $v_\theta$, which is optimized to estimate the velocity vector along the sampling trajectory (see \Cref{subsec:RF}). Similar to SD3 \cite{esser2024scaling}, FLUX.1 replaces the conventional \emph{U-Net} architecture with a fully transformer-based design. A high-level overview of the transformer’s operations at each sampling step is provided in Figure~\ref{fig:transformer} using the notation defined in the official implementation by diffusers \cite{von-platen-etal-2022-diffusers}. On every iteration, the inputs are preprocessed before they ate fed into a sequence of transformer blocks. There are two types of transformer blocks: $(1)$ \emph{Double-Stream} (see \Cref{subsubsec:double}), and $(2)$ \emph{Single-Stream} (see \Cref{subsubsec:single}. Both blocks employ a \emph{Multi-Modal Attention}: joint self-attention mechanism to process concatenated text and image tokens in a unified attention operation, enabling bidirectional interactions that capture both self-information and cross-modal information between text and image modalities. Below we provide a step-by-step guide to the per-iteration pre-process (\Cref{subsubsec:preprocess}), and the \emph{Double-Stream} (\Cref{subsubsec:double}) and \emph{Single-Stream} (\Cref{subsubsec:single}) attention blocks.

\subsubsection{Per-Iteration Pre-process}
\label{subsubsec:preprocess}

\begin{figure}
\centering
\includegraphics[width=0.79\textwidth]{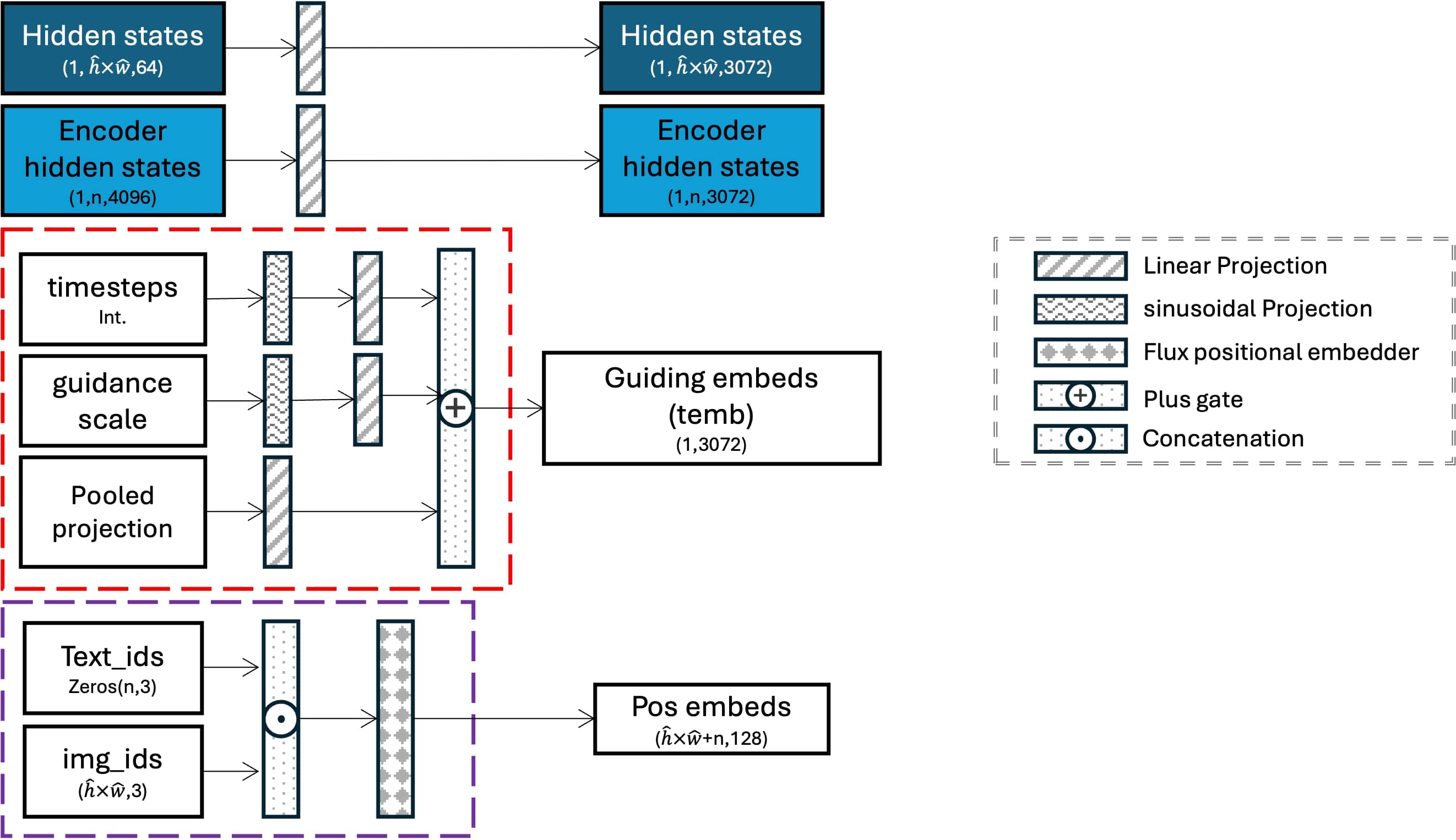}
  \caption{Flux Transformer: per-iteration inputs pre-process}
\label{fig:t_preprocess}
\end{figure}

Along the sampling trajectory, the transformer is called multiple times, once at each sampling iteration. In every call, it is provided with an updated set of parameters: 

\begin{itemize}
    \item the guiding parameter \emph{timestep} is iterated from the pre-calculated list of values between \emph{1} to \emph{0}.
    \item $z_t$ (refers as hidden\_state in the official implementation by diffusers \cite{von-platen-etal-2022-diffusers}) is the latent representation which is (iteratively) refined from Gaussian noise into a clear image. It is being updated along the sampling trajectory.
\end{itemize}

These parameters join the ``constant" inputs (which remain constant across different iterations) pre-calcualted in the pipeline's pre-process stage (\Cref{subsubsec:init}). The transformer pre-processes it's inputs internally as described in \Cref{fig:t_preprocess}. Formally, the following steps are taken:

\begin{itemize}
    \item use of per-domain linear layers to bring the latent embeddings and dense-prompt-embeddings (T5) into a shared dimensionality of 3072 features per token. 
    \item construction of guiding embeds. Unlike Diffusion Models, that encode only the ``temporal" information (i.e, timestep) in this phase, FLUX.1 uses both the timestep and the pooled prompt embeds (from CLIP) as guiding embeddings, yet keeps the traditional notation \emph{temb}. It uses sinusoidal projection (As is \cite{ho2020denoising}) to embed the integer values of the timestep and guidance, than applies dedicated linear projection layers to each component to bring them to the shared dimensionality of 3072 features. Finally, the 3 projected embeddings are summed to create the final \emph{temb} (marked in a red block in \Cref{fig:t_preprocess})
    \item the img\_ids and text\_ids (see \Cref{subsec:arch}) are concatenated, than used to extract per-token positional embeds. To extract positional embeddings from 3D token indices (t,h,w), each axis is first converted to a continuous embedding using axis-specific sinusoidal frequencies scaled by a constant $\theta$. The cosine and sine of the resulting frequency-position products are computed and interleaved to form real-valued vectors. These are then concatenated across all axes to produce the final positional embedding as a pair of tensors (cosine and sine), ready to be used in rotary positional encoding (the process os composed in the purple block in \Cref{fig:t_preprocess}). Note that this process is not influenced neither by \emph{timestep} nor \emph{hidden\_states}, hence the pos\_embeds are constant across different sampling steps.
\end{itemize}  
The \emph{pos\_embeds} and \emph{temb} parameters are used to support the attention mechanism along the step, while \emph{hidden\_states} and \emph{encoder\_hidden\_states} are being processed and refined along the step. \\

\subsubsection{Double-Stream Transformer Block}
\label{subsubsec:double}

\begin{figure}
    \centering
    
    \begin{subfigure}{0.9\linewidth}
        \centering
        \includegraphics[width=\linewidth]{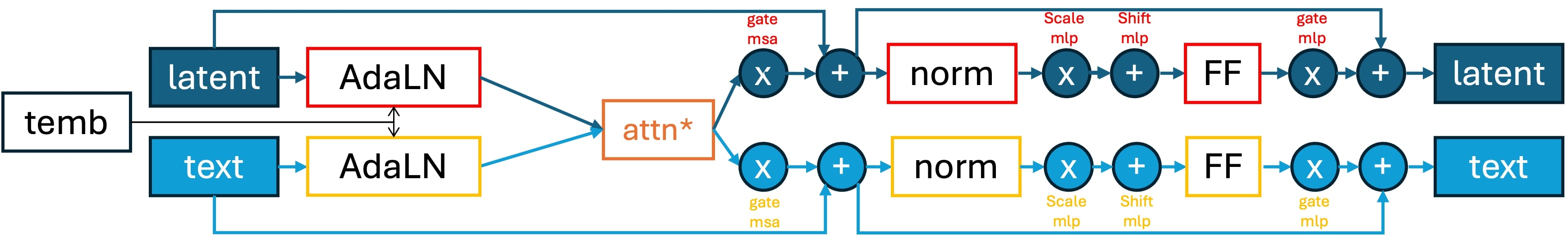}
        \caption{Double-Stream Attention Block: latent and prompt embeddings are processed separately, in a standard multi-modal attention scheme.}
        \label{subfig:double}
    \end{subfigure}
    
    \vspace{1em} % vertical space between subfigures
    
    \begin{subfigure}{0.9\linewidth}
        \centering
        \includegraphics[width=\linewidth]{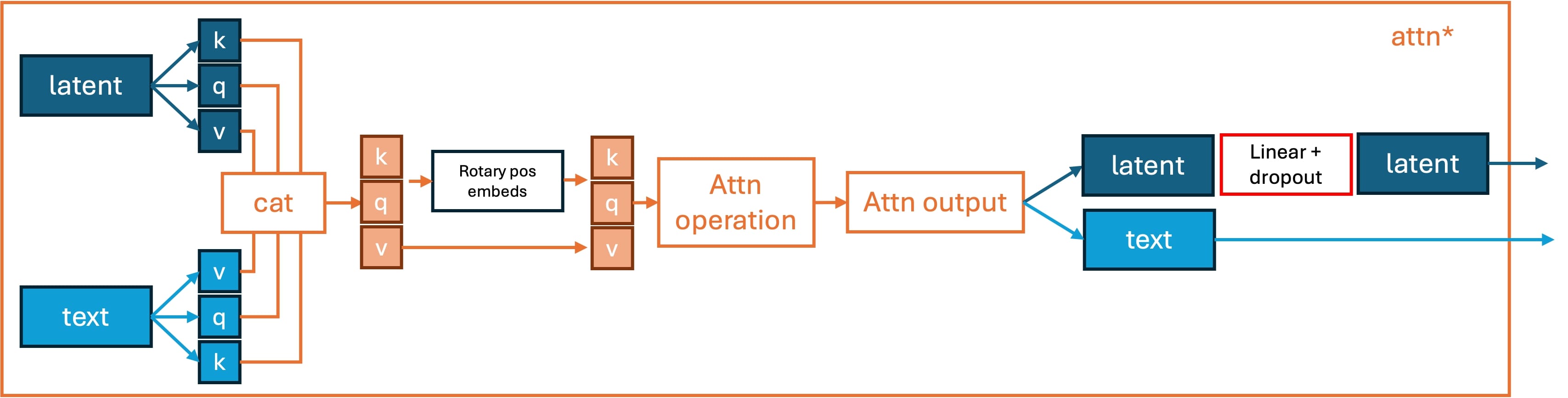}
        \caption{The attention operation is applied over the concatenation of the tokens.}
        \label{subfig:double_Attn}
    \end{subfigure}
    
    \caption{Double-Stream Block.}
    \label{fig:Double_block}
\end{figure}

After pre-process, a series of \emph{19} Double-Stream Transformer blocks are applied. Those blocks employ separate weights for image and text tokens. Multi-modality is achieved by applying the attention operation over the concatenation of the tokens (see \Cref{subfig:double_Attn}). 

\begin{figure}
\centering
\includegraphics[width=0.89\textwidth]{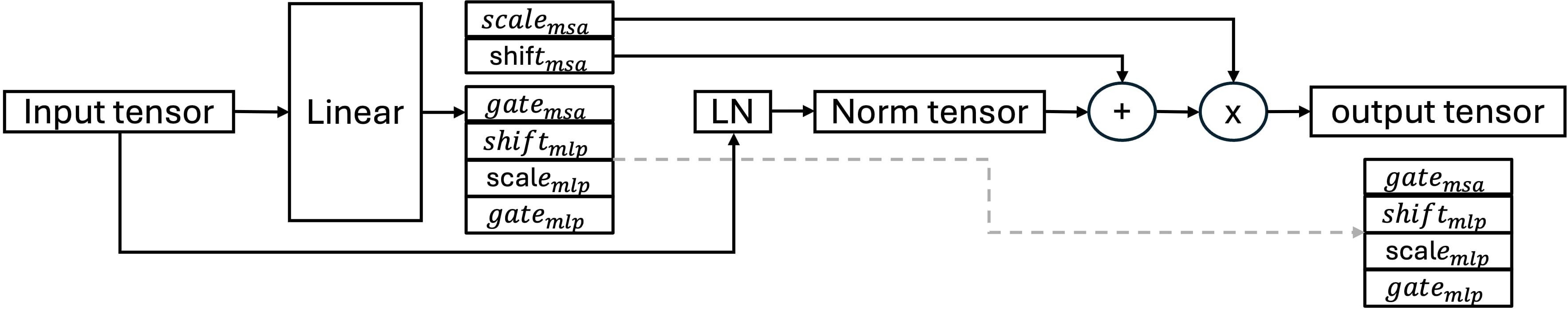}
  \caption{AdaLN layer, where MSA (Multi-head Self Attention) and MLP (Multi-Layer Processor) modulation parameters are computed based on the input tensor. In Single-Stream block (see \Cref{subsubsec:single}), MLP modulation is not computed.}
\label{fig:adaln}
\end{figure}

Each stream (latent and prompt) uses  Adaptive Layer Normalization (AdaLN) \cite{keddous2024vision} layers for normalization and modulation (see \Cref{fig:adaln} and \Cref{subfig:double}). AdaLN is a conditioning mechanism used in Transformer-based models \cite{nichol2021glide, sauer2023stylegan} to modulate intermediate activations based on external input, such as text or image embeddings, based on the Adaptive Instance Normalization (AdaIN) mechanism proposed in \cite{huang2017arbitrary}. Unlike standard Layer Normalization, which applies fixed scaling and shifting parameters, AdaLN dynamically generates these parameters as functions of a conditioning vector (see \Cref{fig:adaln}). This allows the model to adapt its behavior at each layer according to the input prompt or guidance signal.

The normalized tensors are used to extract the $K$, $Q$, and $V$ matrices for each domain, which are then concatenated for mixed attention. The concatenated $K$ and $Q$ matrices are rotated using precomputed Rotary Positional Embeddings. Rotary Positional Embeddings (RoPE) \cite{su2024roformer} are a method for injecting positional information into Transformer models by rotating the query and key vectors in multi-head self-attention according to their token positions. Unlike traditional sinusoidal or learned positional embeddings that are \textbf{added} to input tokens, RoPE applies a \textbf{rotation} in the complex plane that preserves relative positional relationships across sequences. This approach allows the model to generalize better to unseen sequence lengths and supports extrapolation beyond training data. RoPE has become popular in recent vision-language models, where understanding spatial relationships between tokens, especially when re-purposing textual positions for image patches, is crucial.

The rotated $K$ and $Q$ matrices are passed through the mixed attention operation. The outputs are separated back into their respective streams and alpha-blended with the residual input using the $gate_{\text{msa}}$ parameter, extracted by the AdaLN layer. The results are further normalized via a LayerNorm ($LN$) layer and modulated using the $scale_{\text{mlp}}$ and $shift_{\text{mlp}}$ parameters, also extracted by the AdaLN layer, before being passed to the feedforward layer. Finally, the output is alpha-blended with the unnormalized input using the $gate_{\text{mlp}}$ parameter from the AdaLN layer. The entire process is illustrated in \Cref{subfig:double}. \\

\subsubsection{Single-Stream Transformer Block}
\label{subsubsec:single}
\begin{figure}
    \centering
    
    \begin{subfigure}{0.9\linewidth}
        \centering
        \includegraphics[width=\linewidth]{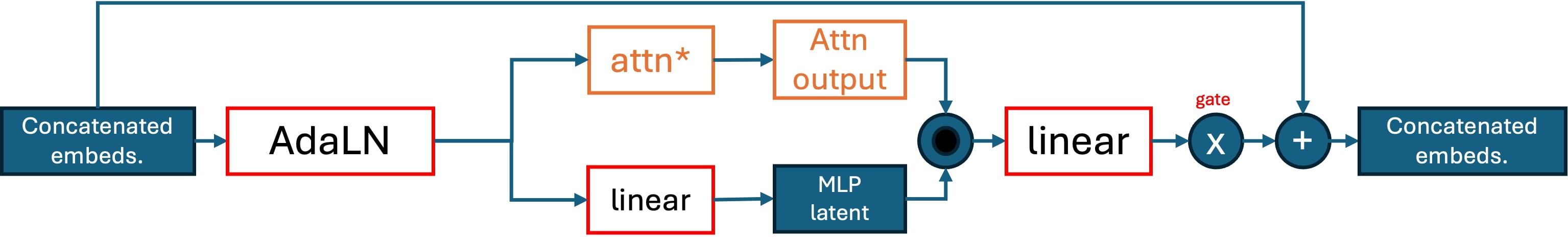}
        \caption{Single-Stream Attention Block: latent and prompt embeddings are processed together, in a simultaneous attention scheme.}
        \label{subfig:single}
    \end{subfigure}
    
    \vspace{1em} % vertical space between subfigures
    
    \begin{subfigure}{0.9\linewidth}
        \centering
        \includegraphics[width=\linewidth]{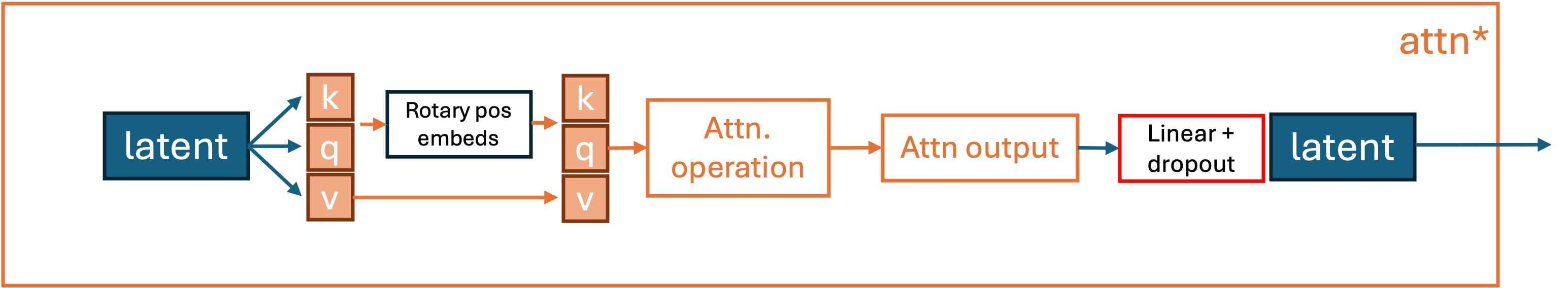}
        \caption{The $K$, $Q$ and $V$ metrices are computed directly from the concatenated representation.}
        \label{subfig:single_Attn}
    \end{subfigure}
    
    \caption{Single-Stream Block.}
    \label{fig:Single_block}
\end{figure}

After the series of Double-Stream blocks is applied, the processed latent and prompt embeddings are concatenated and fed through a series of Single-Stream blocks.
While the Double-Stream blocks apply different weights for prompt and latent embeddings, the Single-Stream blocks use a single set of weights to process a concatenated tensor or latent and text embeddings (see \Cref{fig:Single_block}). In addition, the Single-Stream blocks replace the standard sequential attention block (where an MLP, like Feed-Forward is applied after the attention step), with a parallel mechanism where the attention block and an MLP are computed simultaneously from the same input (see \Cref{subfig:single}). This should not be confused with the multi-modal attention mechanism, which is used in both Double-Stream and Single-Stream blocks, where self-attention and cross-attention are computed jointly in parallel.

A comparison between the Double-Stream and Single-Stream block is summarized in \Cref{tab:stream_blocks_comparison}:

\begin{table}[htbp]
\centering
\begin{tabular}{|l|p{5.2cm}|p{5.2cm}|}
\hline
\textbf{Property} & \textbf{Double-Stream Block} & \textbf{Single-Stream Block} \\
\hline
\textbf{Weight Sharing} & Uses separate weights for text and latent tokens in both attention and feedforward layers. & Uses shared weights for both text and latent tokens across attention and feedforward layers. \\
\hline
\textbf{Computation Style} & Attention and feedforward (MLP) layers are applied sequentially, where the attention's output defines the feedforward's input. & Attention and feedforward layers are computed in parallel using the same input for both. \\
\hline
\end{tabular}
\caption{Comparison between Double-Stream and Single-Stream blocks in FLUX.1}
\label{tab:stream_blocks_comparison}
\end{table}

Specifically, the Double-Stream and Single-Stream blocks differ in two main attributes: weight sharing between text and latent tokens (shared vs. not shared), and computation style (sequential vs. parallel). While the exact motivation behind the FLUX.1 developers choice of this specific architecture is not documented, the following is a brief comparison of these attributes, highlighting the possible pros and cons of each, potentially shedding light on the reasoning behind including both block types in FLUX.1's architecture. \\

\noindent\textbf{Weight Sharing (Shared vs. Not Shared)} 

\noindent Weight sharing refers to whether the same attention and feedforward parameters are used for both text and latent tokens. In the Single-Stream block, shared weights enable more efficient parameter usage and promote tighter integration between modalities, which may help the model generalize better across domains. However, this comes at the cost of reduced flexibility, as both token types must be processed identically despite potentially having very different characteristics. In contrast, the Double-Stream block avoids weight sharing, allowing the model to specialize its representations for text and latent tokens independently. This specialization can improve performance when domain-specific distinctions are critical, though it increases model size and computational overhead. The inclusion of both strategies in FLUX.1 suggests a deliberate balance between efficiency and specialization. \\

\noindent\textbf{Computation Style (Sequential vs. Parallel)} \par

\noindent The key distinction in computation style lies in how the attention and feedforward (MLP) layers are applied within a block. In the Double-Stream block, the computation is sequential: the input is first normalized, passed through an attention layer, then normalized again before being fed into the feedforward layer. This means the output of the attention block defines the input to the MLP feedforward, enabling tightly coupled, stage-wise processing that allows each layer to build upon the previous one. In contrast, the Single-Stream block follows a parallel design. The input is normalized once, and the resulting representation is simultaneously passed through both the attention and MLP layers independently. Their outputs are then combined downstream. This parallelism increases efficiency and allows for broader representation capacity per block, but it may limit the depth of inter-layer interaction present in sequential designs. FLUX.1's use of both may reflect a trade-off between expressive sequential processing and the speed or simplicity of parallel computation. \\

\noindent In summary, Single-Stream blocks emphasize efficiency and simplicity through parallel computation and shared weights, while Double-Stream blocks favor specialization and expressiveness with sequential flow and separate weights. While the Double-Stream blocks follow the \emph{mm-DiT} design previously used in \emph{SD3} \cite{esser2024scaling}, the addition of Single-Stream blocks may reflect the FLUX authors’ intention to expand the model’s capacity in a relatively lightweight and efficient manner.
\section{Models and Tools}
\label{sec:hub}

\subsection{text-to-image}

The FLUX.1 suite of text-to-image models comes in three variants, all share the same architecture of 12B parameters, striking a balance between accessibility and model capabilities \cite{FLUXAnnounce}.

\begin{itemize}
    \item \textbf{FLUX.1[pro]} This the highest-performing variant of the FLUX.1 family, offering superior prompt alignment, visual detail, and output diversity. It is designed for commercial use and is accessible only through licensed API endpoints such as Replicate or Fal.ai. The model weights are not publicly released, making it a hosted-only solution. This version is ideal for production environments where top-tier image generation quality is critical and commercial licensing is feasible.
    \item \textbf{FLUX.1[dev]} provides an open-weight alternative to Pro, guidance-distilled \cite{meng2023distillation} from it. It is licensed strictly for non-commercial use, making it accessible to researchers, developers, and hobbyists for experimentation and academic work. Unlike Pro, Dev can be downloaded and run locally, offering full control and flexibility for non-commercial applications. The FLUX.1 [dev] model is available via huggingface can be directly tried out on Replicate or fal.ai. It should be noted that the Dev version might require more sampling steps than the pro version to achieve high-level performance.
    \item \textbf{FLUX.1[schnell]} is a speed-optimized, distilled variant of FLUX.1 Pro, designed to significantly reduce inference time while retaining reasonable image quality. It shares the same 12B architecture but is trained via timesteps-distillation \cite{sauer2024fast} from the Pro model, meaning it learns to mimic Pro’s outputs using fewer sampling steps, up to a single step, by optimizing for speed and efficiency. This results in faster generations at the cost of some prompt fidelity and fine detail. Released under the permissive Apache 2.0 license, Schnell allows unrestricted commercial use, making it ideal for real-time, low-latency applications and lightweight deployments where speed and open licensing outweigh the need for peak visual fidelity.
    \item \textbf{FLUX1.1[pro]} is an enhanced version of the FLUX.1[pro] model, capable of faster image generation while also improving image quality, prompt adherence, and diversity. The model introduces new modes—Ultra, enabling 4x higher resolution without compromising speed, and Raw, producing hyper-realistic, candid-style images. Additionally, a prompt upsampling feature leverages large language models (LLMs) to expand and enrich user prompts, enhancing creative outputs. While not covered in this report, FLUX 1.1 Pro is available via API on platforms like Replicate and Fal.ai, with commercial licensing required.

\end{itemize}

\subsection{tools}

Originally developed for text-to-image generation, FLUX.1 has since been extended by Black Forest Labs with a suite of tools supporting a range of use cases beyond its original purpose. This report does not explore each tool in depth as it does for the FLUX.1 model itself. Below is a list of publicly available tools, current as of June 5th, 2025:

\begin{itemize}
    \item \textbf{FLUX-Fill} is an inpainting and outpainting models, enabling editing and expansion of real and generated images given a text description and a binary mask.
    \item \textbf{FLUX-Canny} enables structural guidance based on canny edges extracted from an input image and a text prompt.
    \item \textbf{FLUX-Depth} enables structural guidance based on a depth map extracted from an input image and a text prompt.
    \item \textbf{FLUX-Redux} is an adaptor that aligns dense (per-token) image embeddings extracted by SigLIP image encoder \cite{zhai2023sigmoid} with the T5 text-embedding's space for image-conditioned generation with a pre-traine FLUX.1 base model. While limited for image variation in the Dev version, It integrates into more complex workflows unlocking image restyling via prompt when using the FLUX1.1 [pro] Ultra model, allowing for combining input images and text prompts.
    \item \textbf{FLUX-Kontext} extends the text-to-image capabilities of the base FLUX model into powerful in-context generation and editing workflows by enabling multimodal conditioning—allowing users to guide generation using both text and reference images. While \emph{FLUX-Redux} had already introduced basic multimodal conditioning for image variation and prompt-driven restyling, \emph{FLUX-Kontext} unlocks advanced functionality such as localized edits, style transfer, and character or scene consistency across images. With models like Kontext [pro] and [max], the focus shifts toward fast, iterative generation and high-precision editing. As of June 5th, 2025, the \emph{FLUX-Kontext} models are not yet publicly released, and therefore could not be directly explored or evaluated in this report.

\end{itemize}

\phantomsection
\addcontentsline{toc}{section}{References}
\bibliographystyle{IEEEtran}
\bibliography{main}

\newpage
\setcounter{table}{0}
\setcounter{figure}{0}
\appendix
\appendixpage
\addappheadtotoc
\section{Qualitative Evaluation}
\label{Appendix}
In this section, we provide a qualitative comparison of FLUX.1's performance against Stable Diffusion (SD) 2.1, which remains widely used in various data generation projects across R\&D and Engineering. We use FLUX.1\emph{[dev]} with 50 sampling steps and a guidance scale of 2.5. For SD 2.1, we use 50 sampling steps and a guidance scale of 7.5. All comparisons are conducted at a resolution of $512 \times 512$, which is the recommended resolution for SD 2.1. We use the publicly available version of the two models without any additional finetuning. 

\begin{figure}[p]
\centering
\includegraphics[width=0.99\textwidth]{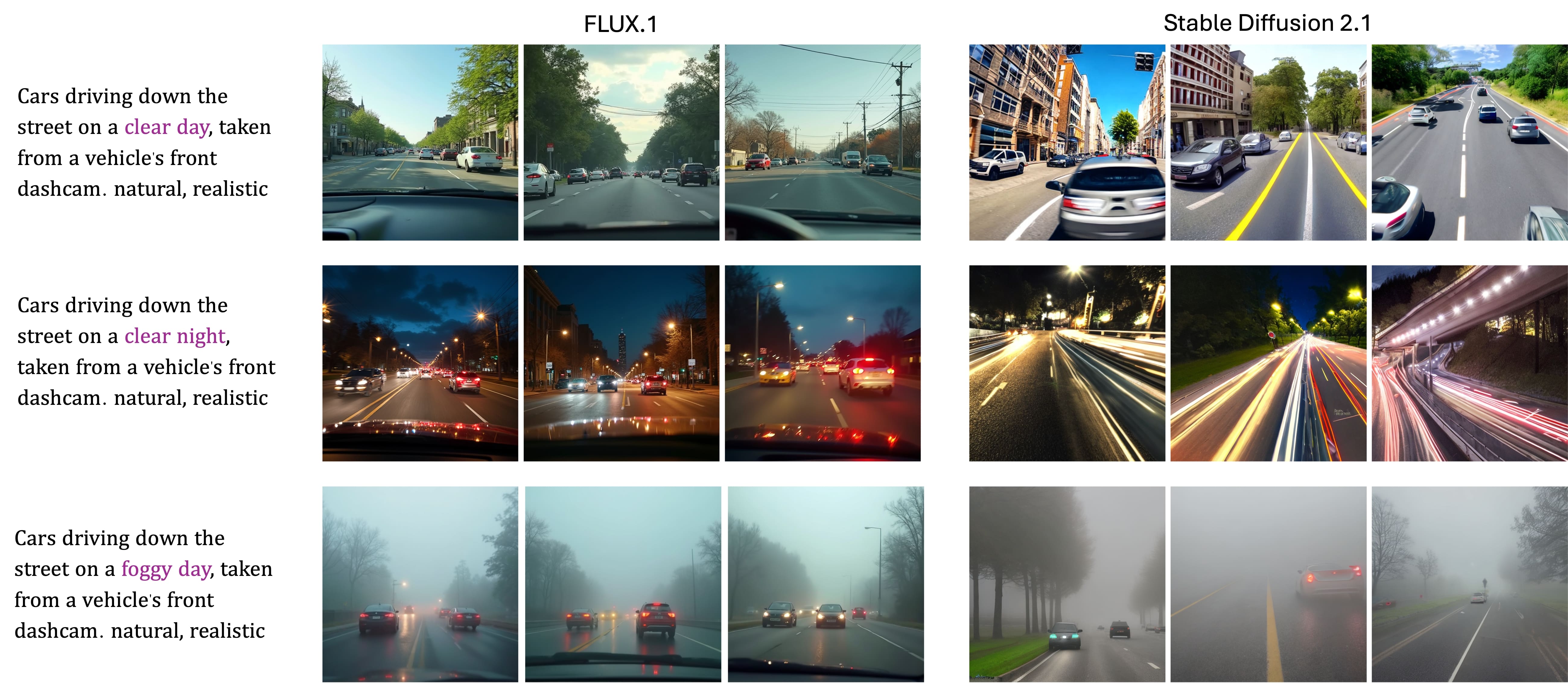}
  \caption{Overall image quality comparison. FLUX.1 generates high quality automotive scenes with different attributes (here: weather and lighting) without additional optimization.}
\label{fig:comp_quality}
\end{figure}

\begin{figure}[p]
\centering
\includegraphics[width=0.99\textwidth]{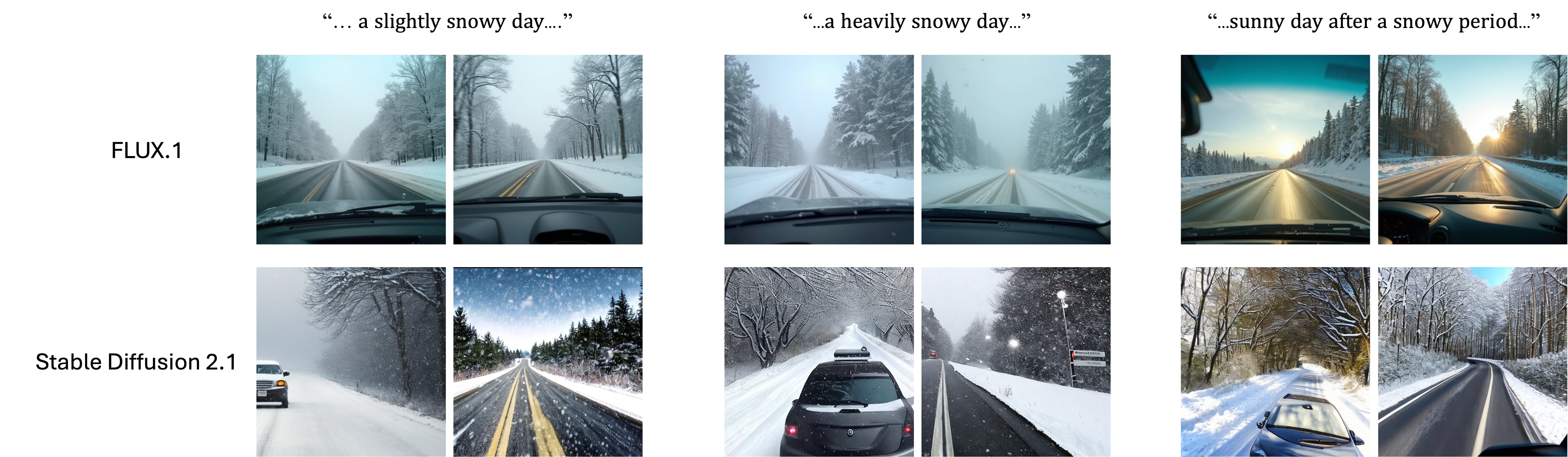}
  \caption{FLUX.1 is sensitive to subtle differences in textual guidance- for example, distinguishing between light and heavy snow. It also handles complex attributes, such as in the prompt: ``A sunny day after a snowy period''.}
\label{fig:comp_adherence}
\end{figure}

\begin{figure}[p]
\centering
\includegraphics[width=0.99\textwidth]{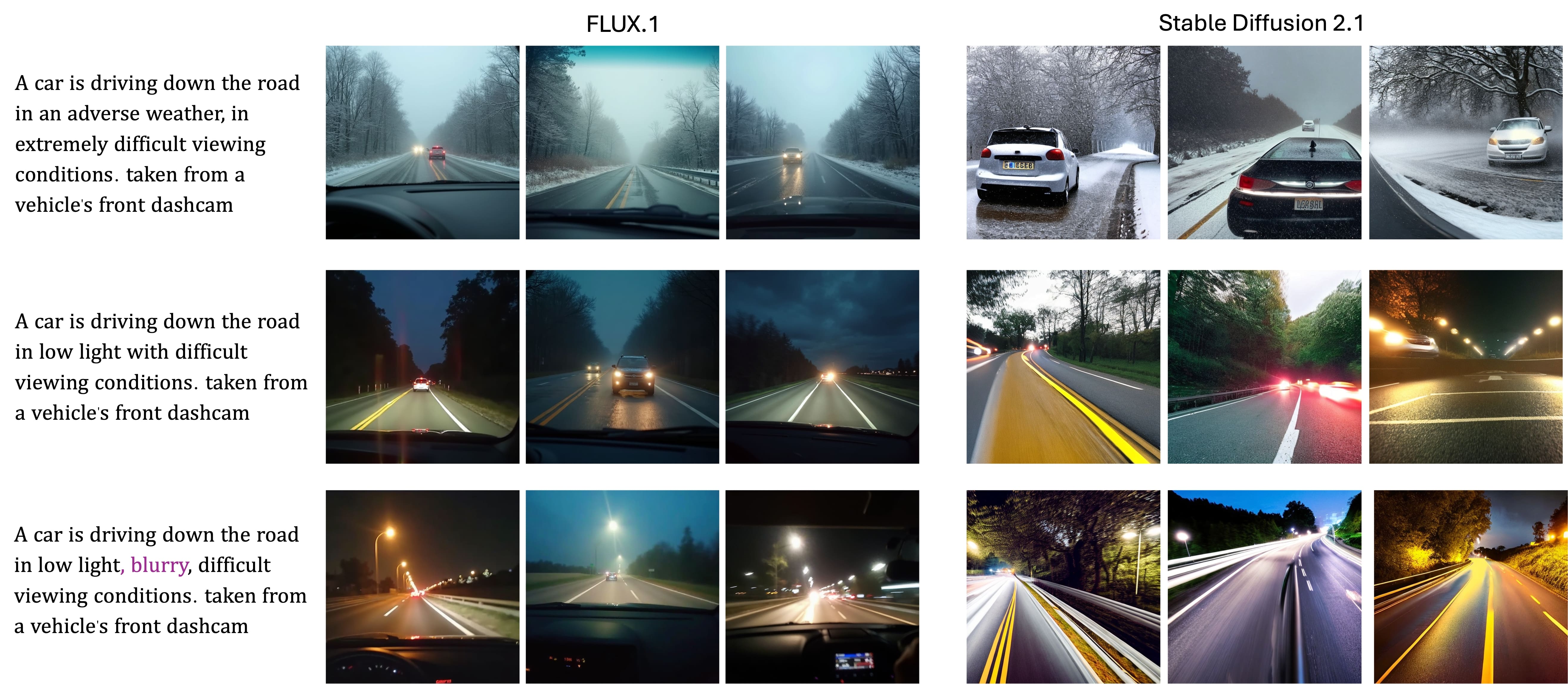}
  \caption{FLUX.1 is capable of interpreting textual guidance that is minimal or loosely defined, such as `adverse weather'' ``difficult viewing conditions'' while remaining sensitive to direct requests like ``blurry''.}
\label{fig:comp_sparse_prompt}
\end{figure}

\begin{figure}[p]
\centering
\includegraphics[width=0.99\textwidth]{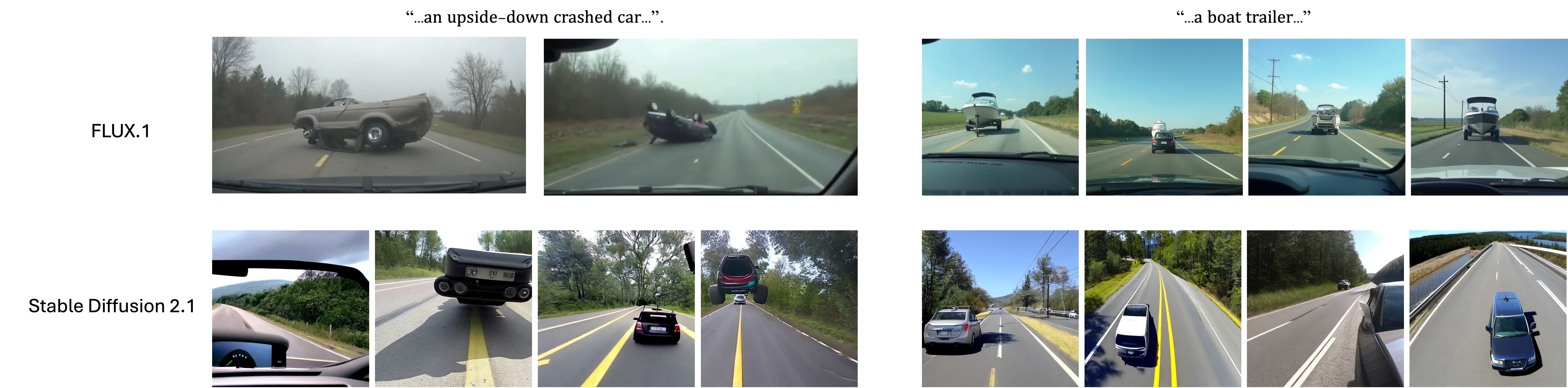}
  \caption{FLUX.1's rich vocabulary and extensive pre-training enable it to handle rare concepts such as ``boat-trailer'' and ``upside-down crashed car''.}
\label{fig:comp_rare_concepts}
\end{figure}

\begin{figure}[p]
\centering
\includegraphics[width=0.99\textwidth]{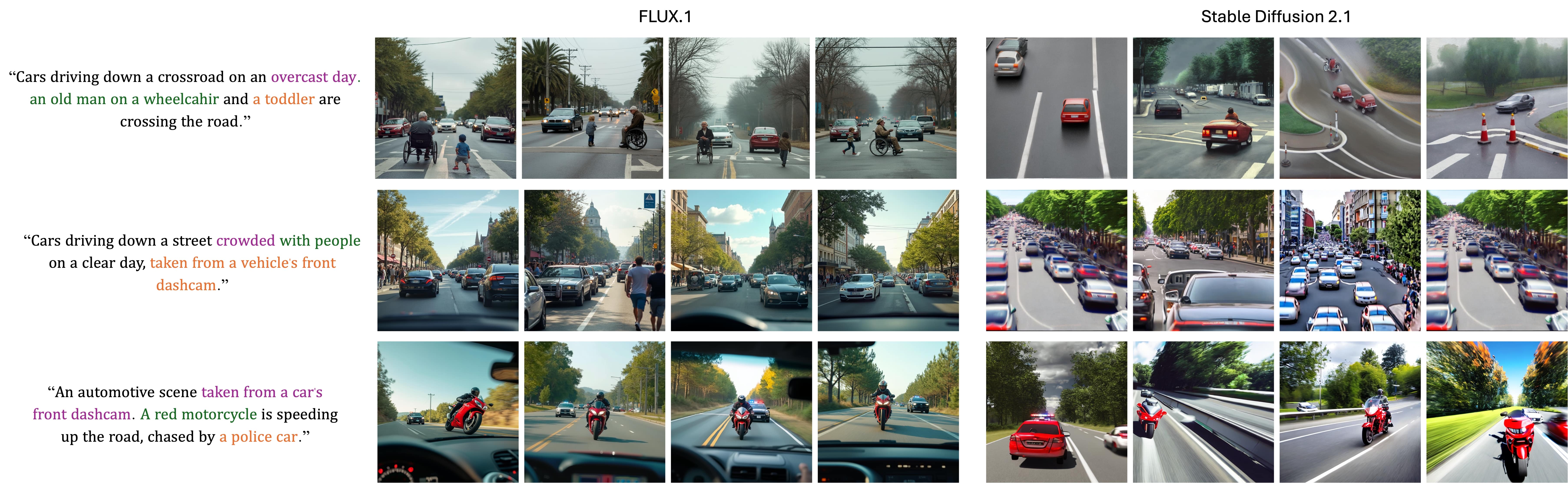}
  \caption{The dense prompt embeddings produced by the T5 text encoder enable FLUX.1 to adhere to complex prompts containing multiple requests.}
\label{fig:comp_complexPrompt}
\end{figure}

We compare the models' performance on text-to-image generation of automotive scenarios across the following aspects:

\begin{itemize}
    \item Overall image quality (\Cref{fig:comp_quality})
    \item text-image alignment (\Cref{fig:comp_adherence})
    \item Alignment to sparse prompts: text descriptions that are minimal or loosely defined. (\Cref{fig:comp_sparse_prompt})
    \item Synthesis of rare concepts (\Cref{fig:comp_rare_concepts})
    \item Adherence to complex prompts containing multiple descriptive elements. (\Cref{fig:comp_complexPrompt})
\end{itemize}

In all examples, even if not specifically noted, the models were prompted to generate images as if captured from a vehicle's front dashcam. While SD 2.1 struggled to maintain the specifically requested viewing angle, FLUX.1 consistently remained faithful to it across different tasks.

\begin{figure}[p]
\centering
\includegraphics[width=0.89\textwidth]{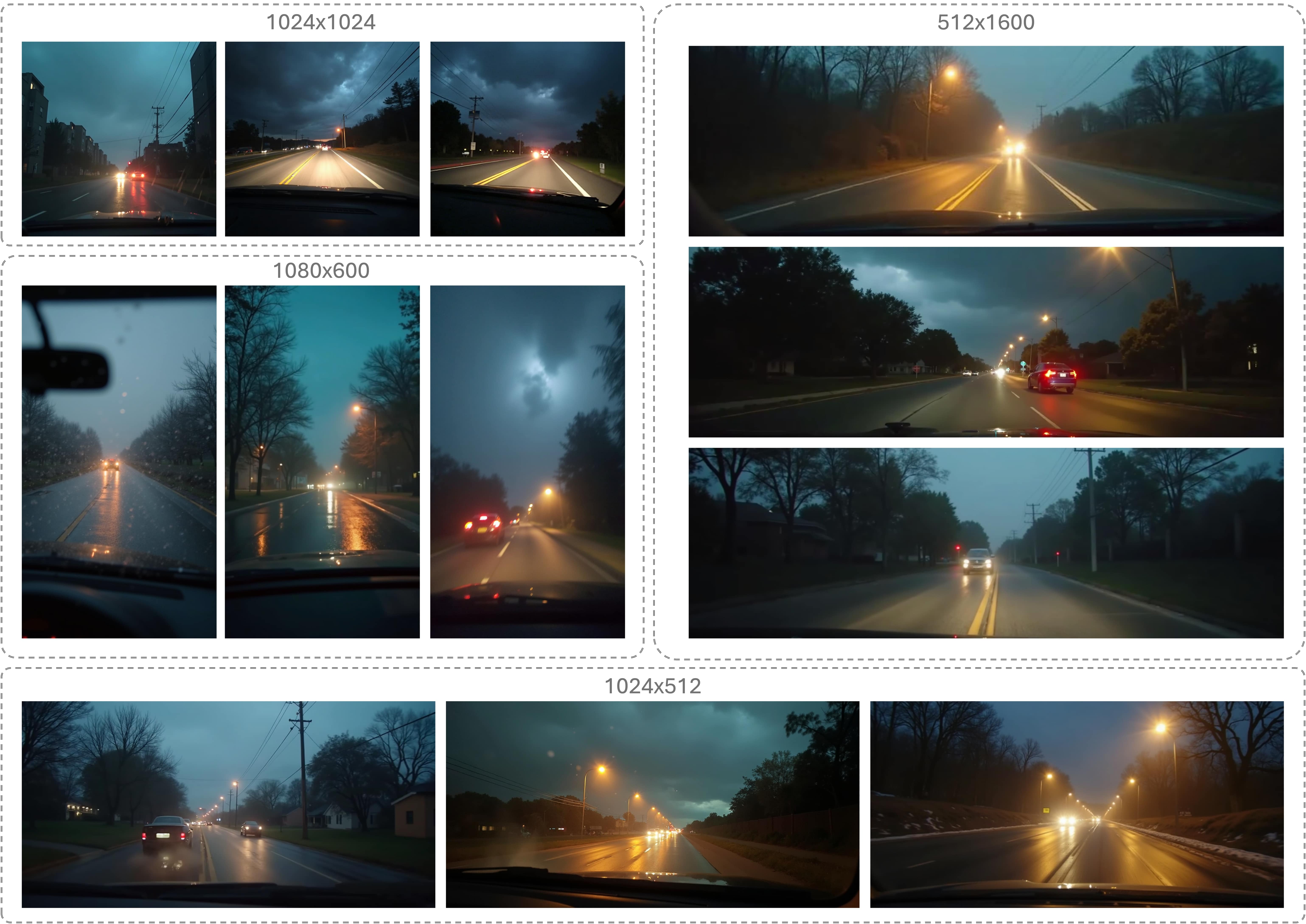}
  \caption{While existing U-Net–based diffusion models are typically limited to a specific or fixed set of resolutions, FLUX.1 can operate across a wide range of resolutions and aspect ratios.}
\label{fig:resolutions}
\end{figure}

We also demonstrate FLUX.1 capability to generate images in various resolutions and aspect ratios (\Cref{fig:resolutions}).

All in all, FLUX.1 brings state-of-the-art genrative capabilities in all aformentioned aspects, compared to previous models, and particularly compared to SD2.1. These gaps become more pronaunced when relates to complex generative tasks, such as the automotive case, where FLUX's rich voucabulry and heavy pre-training enable hin to deal with generative tasks that are way out of distribution for previos models. It also handels high-resolution images with various aspect ratios. These capabilities may become handy in automotive-related augmentation tasks like Adverse viewing conditions and rare-concpts generation/inpaining, as well as in data restoration and enhancement.

\end{document}